# Rapport du Projet de Recherche TRAIMA

TRaitement Automatique des Interactions Multimodales
en Apprentissage

MSHS - Thématique « Apprentissage et numérique »

**Responsable du projet**
Jean-François CERISIER
Professeur des Universités
Vice-président de l'université de Poitiers
et de l'université confédérale Léonard de Vinci
Directeur du laboratoire TECHNE EA 6316
Titulaire de la chaire régionale MORTIMER

**Laboratoires concernés**
Laboratoire TECHNE EA6316
Laboratoire FORELLIS EA3816
Laboratoire LIAS, ISAE ENSMA et Université de Poitiers EA 6315

**Liste des personnes impliquées par laboratoire**
*Laboratoire TECHNE EA6316 :*
Jean-François Cerisier
Emilie Remond
Aurélien Nguyen
Andrew Peterson
*Laboratoire FORELLIS EA 3816 :*
Julie Rançon
*Laboratoire LIAS EA 6315 :*
Ladjel Bellatreche

Projet mené de mars 2019 à juin 2020

**Résumé du projet**

Le projet de recherche TRAIMA a pour objectif d'explorer le potentiel de traitement automatique des interactions multimodales en situations pédagogiques. Actuellement, l'analyse des interactions verbales, paraverbales et non-verbales est effectuée manuellement par les chercheurs. Très couteuse en temps, elle pourrait être remplacée par des algorithmes de *Machine Learning* qui les catégoriseraient et les classifieraient.

Notre travail a consisté à faire état des avancées scientifiques sur le sujet et définir une méthodologie de travail sur le traitement automatique des interactions explicatives.

**Contexte scientifique**

Le rapport de conjoncture du CNRS (2014 : 159) note que la prise en compte du langage en contexte naturel rend nécessaire le traitement de la multimodalité, intégrant les aspects verbaux et non verbaux de l'interaction (gestes, attitudes, etc.). Les rapporteurs soulignent que la direction des études en TALP (traitement automatique de la langue et de la parole) se doit de porter à présent sur la sémantique pour laquelle il n'existe toujours pas de modèle véritablement utilisable. Il est encore difficile de passer d'une approche de niveau lexical à des représentations structurées de type pragmatique. Ainsi, il devient nécessaire de rapprocher les technologies à base de règles aux méthodes d'apprentissage.

D'un point de vue purement linguistique, les chercheurs en didactique se sont focalisés sur des analyses interactionnelles qualitatives basées sur des observations de tours de parole en classe, de places dans le groupe et la manière dont les participants les occupent (Kerbrat-Orecchioni, 2001 ; Cicurel, 2011). Ces investigations portant sur des micro-analyses d'événements discursifs sont aujourd'hui enrichies d'un intérêt pour les comportements non-verbaux notamment la place des gestes professionnels, du jeu des postures (Bucheton, 2009). Le corps de l'enseignant, au sens de Tellier (2008), est un véritable outil pédagogique et a pour intention d'informer, évaluer et animer. Il peut également expliquer ou tenter de faire comprendre selon Rançon (2011) mais également de collaborer. Ces attitudes enseignantes sont d'autant plus fondamentales pour l'accès au sens qu'interaction et acquisitions en contexte sont reliées (Gajo & Mondada, 2000). Mais au-delà des formes linguistiques et des dynamiques interactionnelles bien identifiées (Mondada, 1995), se pose la difficulté du traitement des corpus.

**Objectifs, enjeux et originalité du projet**

Notre objectif a été principalement d'explorer le potentiel de traitement automatique des corpus interactionnels intégrant la multimodalité (posturo-mimo-gestualité « PMG » des interactants) lors de situations pédagogiques. Plus spécifiquement, nous avons traité en priorité les séquences explicatives (Fasel Lauzon, 2012, 2014) et collaboratives (Gracia-Moreno, 2015, 2016, 2017) en interaction qui ont lieu entre enseignants et apprenants de langue (Français Langue Etrangère et Français Langue Maternelle). L'enjeu a été d'arriver à mettre au point une méthodologie de recherche basée sur des algorithmes de *Machine Learning* afin de catégoriser et classifier les interventions verbales et non-verbales des acteurs pédagogiques. L'originalité de ce projet est qu'il n'existe pas, à ce jour, une application permettant d'automatiser à grande échelle les corpus interactionnels lorsque les intervenants parlent en même temps et pour lesquels la PMG n'est pas forcément signifiante ou vue de tous.

Ce projet s'inscrit notamment dans le contexte de la plateforme TechnéLAB qui se compose d'une salle d'activité avec des mobiliers modulaires et d'une régie équipée pour l'enregistrement audiovisuel dans d'excellentes conditions techniques (captation multicaméras HD, micros HF individuels et micros directionnels, enregistrement séparé et synchronisé des flux audiovisuels, sauvegarde sécurisée et redondante dans deux salles d'hébergement de l'université) mais aussi la captation et l'enregistrement des traces d'interactions numériques, de données de géolocalisation fine et d'attention visuelle (*eye tracking*). Le développement continu de cette plateforme prévoit l'automatisation la plus importante possible de toutes les tâches des chaines de collecte et de traitement de données. Pour ce projet, le TechnéLAB est à la fois le plateau technique qui permettra de collecter les données dont développer nos méthodes et techniques d'analyse automatique des interactions multimodales et la plateforme qui bénéficiera de ces outils au service d'autres programmes de recherche sur les usages des techniques numériques dans le champ de l'éducation.

# Plan du rapport de recherche

# Le traitement automatique des données multimodales :
## Le cas du discours explicatif

Julie RANCON
FORELLIS (EA 3816)

---

Notre objectif est d'explorer le potentiel de traitement automatique des corpus interactionnels intégrant la multimodalité (posturo-mimo-gestualité « PMG » des interactants) lors de situations pédagogiques. Plus spécifiquement, nous avons traité en priorité les séquences explicatives (Fasel Lauzon, 2012, 2014) en interaction qui ont lieu entre enseignants et apprenants de langue (Français Langue Etrangère et Français Langue Maternelle).
Une première étape essentielle du travail a été de définir linguistiquement ce qu'est une séquence explicative en classe afin de constituer un corpus représentatif de ce phénomène.

### 1. La polysémie d' « Expliquer »

La notion d'explication renvoie à un phénomène discursif (on explique quelque chose à quelqu'un) mais aussi à un processus cognitif (on cherche des explications aux phénomènes qui nous entourent, c'est-à-dire, qu'on cherche à leur attribuer une signification). Fasel Lauzon (2009) nous rappelle qu'en latin, expliquer provient de *plicare* (plier), qui renvoie à l'origine à l'idée de quelque chose qu'on déplie, déploie, dé(sen)veloppe et que ce faisant, on montre dans sa totalité. Dans son sens figuré, expliquer est faire voir, faire connaitre, faire comprendre en développant, en montrant, exposant. C'est alors dégager une structure, un fonctionnement, une signification qui ne 'saute pas aux yeux'.

Une autre acception d'expliquer est celle de donner des raisons (on explique le motif d'un crime, la cause d'un accident). Lorsque l'objet à expliquer ne peut pas être déplié, c'est-à-dire que son origine, son fonctionnement ou sa signification ne sont pas directement accessibles. L'explication prend ici un caractère hypothétique. Dans ce sens, c'est avant tout interpréter, construire un raisonnement pour tenter d'élucider un mystère.

Enfin, une dernière acception renvoie à l'explication comme but de la compréhension en soi. Dans certains contextes, la compréhension en vue d'actions qui peuvent être concrètes : comprendre, c'est pouvoir agir. Le sens d'action pour expliquer est alors une explication procédurale (savoir-faire) pour référer à un discours indiquant une démarche étape par étape (règles d'un jeu).

De ces considérations générales, nous pouvons dire que le discours explicatif est une discours de compréhension (De Gaumyn, 1986,1991), qu'il se sépare en l'explanans (ce qui explique) et l'explanadum (ce qui est à expliquer) (Hudelot, 2001) et que le lien causal peut être implicite (Charaudeau & Maingeneau, 2001). Dès lors, l'observation des phénomènes verbaux et non-verbaux s'impose afin de déceler la présence de ce discours de compréhension.

### 2. Différentes formes linguistiques d'explication pédagogique

En 2002, Charaudeau & Maingueneau redéfinissent les catégories de Charaudeau (1992) pour n'en définir plus que trois :

- L'explication causale (qui permet la prédiction)
- L'explication fonctionnelle (« Pourquoi le cœur bat-il ? Pour faire circuler le sang »)
- L'explication intentionnelle (« Il a tué pour voler »).

Dans l'intérêt de notre recherche, l'ensemble de ces formes explicatives ont été prises en considération pour élaborer notre corpus de travail.
Il est à noter que le discours explicatif se trouve à l'orée d'autres types de discours dont les proximités sémantiques sont visibles dans l'interaction, comme présenté dans le schéma suivant :

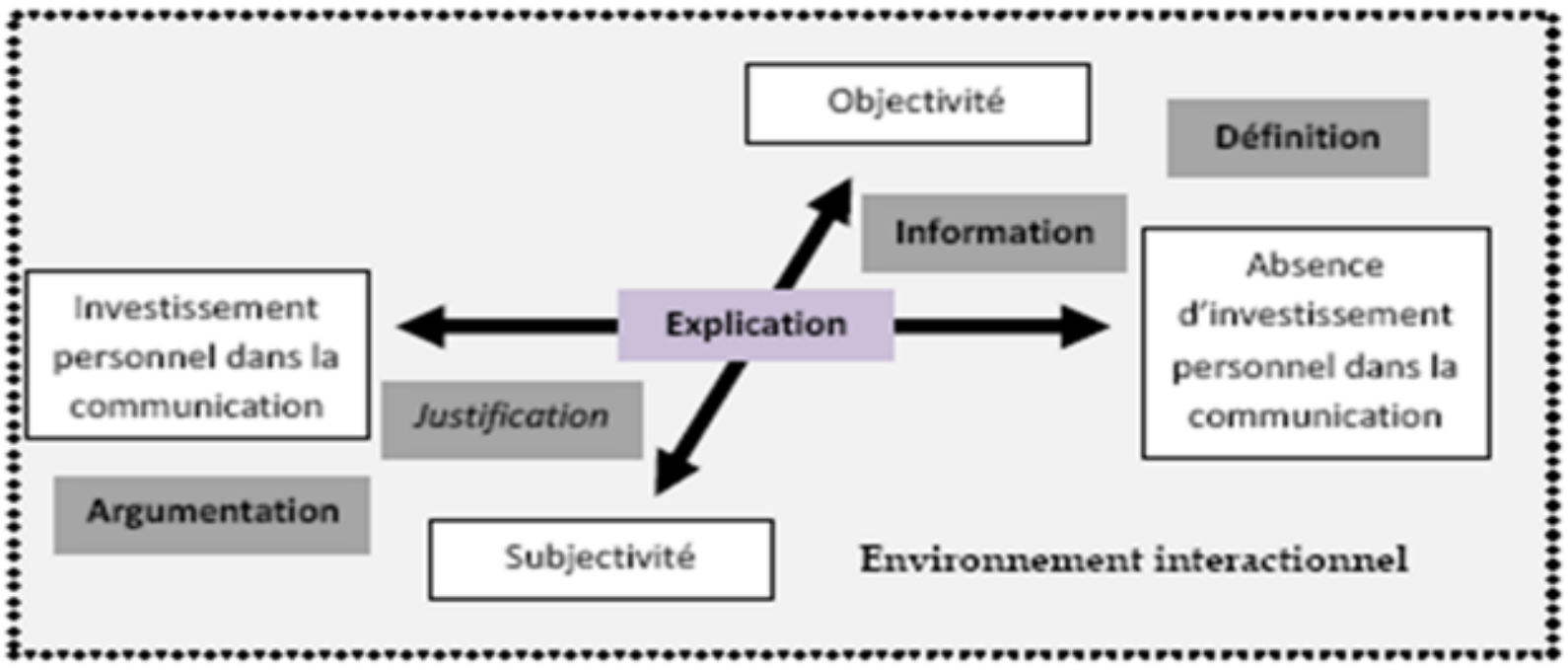

Situations d'existence du discours explicatif

Le discours explicatif peut se trouver dans la position d'un discours définitoire dont les propriétés d'objectivité et d'absence du locuteur dans l'énonciation sont révélatrices. Il peut aussi être perçu comme un discours informatif pour les apprenants à la différence que l'enseignant est attentif aux feed-back de ses élèves.

A l'opposé, le discours explicatif se différencie de l'argumentation dans le sens où faire adhérer le locuteur n'est pas la préoccupation première de l'enseignant. Il est plutôt question de faire comprendre. De même, le discours d'explication est souvent amalgamé avec la justification dont la présence ne semble pas évidente en classe du fait des positions hiérarchiques des locuteurs.

En classe de langue (entendez classe de FLM et classe de FLES), le discours explicatif est aussi un discours de reformulation. L'enseignant est amené à reformuler des informations prononcées en vue de rappeler ce qui vient d'être dit et de créer potentiellement de la signification pour les apprenants. La démarche explicative (Miéville, 1981) organise alors une certaine progression dans l'apprentissage.

En cela, le discours explicatif régule la parole, identifie le moment d'incompréhension mais aussi tente de construire une réalité individuelle (Lund, 2003) (compréhension du lexique par chaque apprenant de la classe) et dans notre cas une réalité textuelle (compréhension du lexique d'un texte). N'oublions pas que le cadre scolaire a pour finalité de faire comprendre mais aussi faire réutiliser les connaissances expliquées (Bogaards, 1994).

## 3. L'explication en interaction comme séquence tripartite

L'explication est un mouvement interactionnel visant à résoudre un problème de compréhension qui constitue un obstacle à la poursuite de l'interaction (Barbieri & al., 1990).

C'est un processus collaboratif de construction négociée de connaissances réalisé par l'action conjointe des participants d'un dialogue et visant l'intercompréhension (Baker, 1992). L'explication est une séquence tripartite :

- Ouverture qui contient la problématisation, la constitution de l'objet à expliquer (explanandum)
- Noyau qui contient la résolution du problème parfois appelée 'l'explication proprement dite' (ou explanans)
- Clôture qui prend la forme d'une marque de réception ou de ratification de l'explication

```
      ((en parlant du personnage principal))
GOU     vraisemblablement  il  n'a  pas  d'travail
        parce qu'il euh: il n'a pas d'quoi manger
        quatre repas
```

| CECI | EXPLIQUE | CELA |
|---|---|---|
| « Vraisemblablement il n'a pas de travail » | [est cause] « parce que » | « il n'a pas de quoi manger 4 repas » |

Causalité de *dicto*: paroles où le lien n'est pas très clair ni précis
Causalité de *re* : présence d'un lien causal réel
(Raccah, 2005)

L'élaboration du message explicatif s'effectue d'une part, en fonction des représentations de l'enseignant sur l'objet à expliquer et d'autre part, du degré d'interprétation de l'apprenant au cours de l'activité cognitive de compréhension du message.

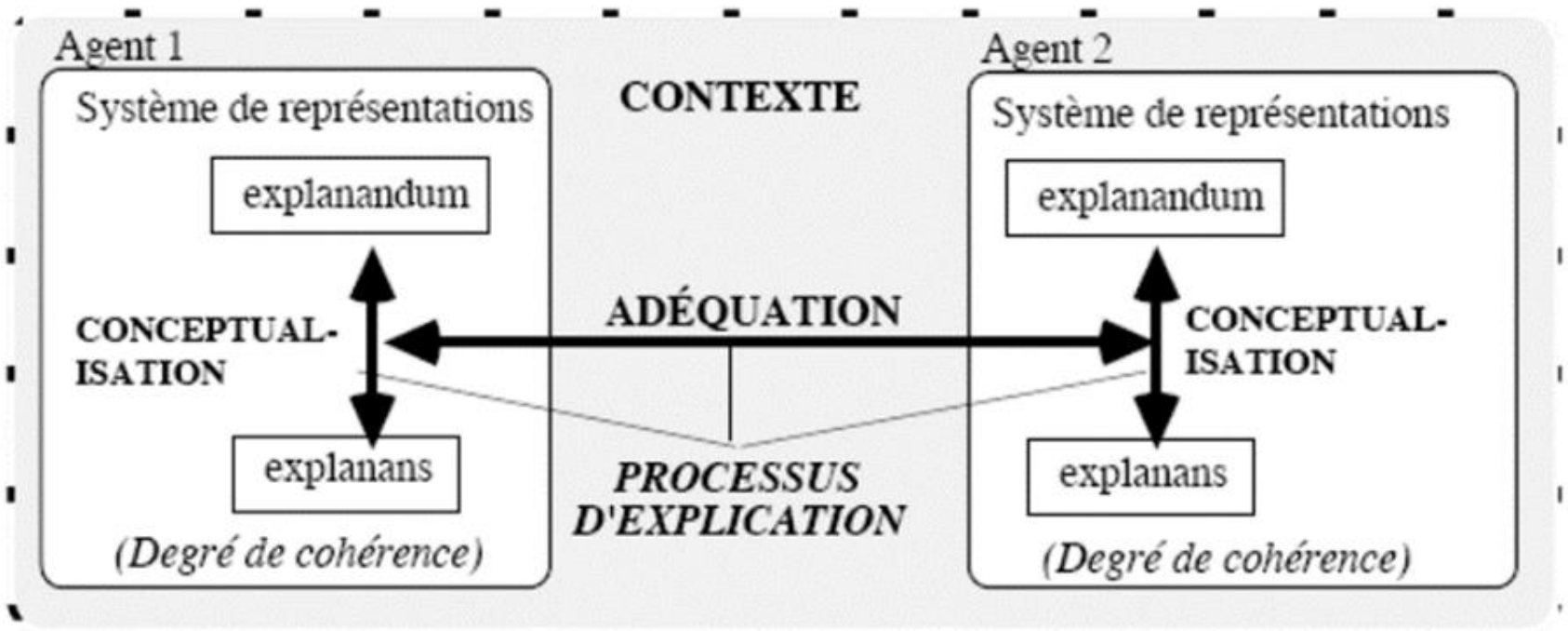

Baker (1994)

Adam (1992) propose en outre un découpage du discours explicatif en analysant une succession de phases : « phase de questionnement + phase résolutive + phase conclusive » en y intercalant la vision de schématisation initiale facultative de Grize (1990 : 107) :

| | | | |
|---|---|---|---|
| 0. | Macro-proposition explicative 0 : schématisation initiale | | |
| 1. | Pourquoi X ? comment X ? | Macro-proposition 1 : problème | (question) |
| 2. | Parce que | Macro-proposition explicative 2 : explication | (réponse) |
| 3. | Macro-proposition explicative 3 : conclusion-évaluation | | |

## 4. Corpus INTER-EXPLIC de séquences explicatives en interaction

Pour les besoins de notre recherche, nous avons sélectionné des séquences explicatives tripartites au sein du corpus INTER-EXPLIC (Université de Toulouse II, 2006), corpus composé d'environ 30h de séquences pédagogiques de FLE et FLM. Plus d'une centaine d'épisodes explicatifs en interaction y ont été recensées (cf. annexe). Nous avons codé manuellement ce corpus à l'aide des conventions de transcription ICOR afin de classifier les discours interactionnels verbaux, paraverbaux et non-verbaux.

Exemple de séquence explicative du corpus INTER-EXPLIC, transcrites manuellement selon les conventions de transcription ICOR :

```
Mic.ilportaitbeau2.mov
Mic.L1ilportaitbeau2.cam2.mov
Il portait beau :

       ((56:32.3))
((les bras sont croisés, la main gauche tient le menton))
MIC    on vous dit qu'il est beau
SIL    (0.9)
ELS    <((réponse inaudible des élèves))(0.6)>
ELE    le beauté c'est pas=
MIC    =<(( les doigts de la main gauche sont rassemblés et montrés aux
       élèves))qu'il porte beau <((les doigts de la main gauche s'ouvrent vers
       l'avant))qu'est-ce que ça signifie/=
ELE    =c'est [bien  ]
THI           [il por]te bien=
MIC    =qu'il le porte bien>
SIL    (0.9)
ELS    <((commentaires inaudibles d'élèves)) (4.6)>
MIC    <((montre le pouce vers son côté gauche))par nature> (0.6) <((montre l'index
       vers sa droite))et par pose> (0.9)ça signifie que <((geste de la main gauche
       et mouvements des doigts))cette beauté elle est (2.1) naturelle> (0.3) non\
SIL    (0.3)
ELS    <((réponse inaudible d'élèves))>=
ELE    =oui
SIL    (0.3)
MIC    oui (0.3) ensuite
       ((56:59.5))
```

Nous avons également un second corpus EXPLIC-LEXIC (Université de Poitiers, 2016) qui aura pour objectif d'être entièrement transcris automatiquement.

## 5. Les explications paraverbales et non-verbales de l'enseignant

Les explications paraverbales utilisent le canal oral pour expliquer, c'est-à-dire toutes les manifestations vocales et prosodiques annexes à une production verbale. Quant aux explications non-verbales, ce sont des manifestations qui utilisent le canal visuel : on y retrouve la kinésique (la posturo-mimo-gestualité), la proxémique (le corps dans l'espace) et les supports iconiques. Pennycook (1985) et Brancroft (1997) in Lazaraton (2004 : 81) suggèrent que les deux tiers de la communication sont portés par le non-verbal et seulement un tiers par le verbal.

**Fonction des explications paraverbales et non-verbales**

Ces techniques peuvent avoir deux fonctions : soit elles accompagnent une explication verbale, soit elles se substituent à elle. Le premier cas est le plus commun car il peut s'appliquer à l'ensemble des lexies à expliquer. Nous n'avons pas observé d'explication paraverbale se substituant totalement à l'explanandum. Pourtant, d'après notre expérience, il est tout à fait possible d'expliquer par l'intermédiaire de ces techniques. Par exemple, pour expliquer ce qu'est un fiacre, l'enseignant peut montrer une image représentant le moyen de locomotion en question. Pour l'heure, nous n'avons observé que des techniques paraverbales/non-verbales accompagnant un contenu explicatif verbal. Ces techniques s'associent également entre elles. Nous déterminerons si cette association quantitative a un effet qualitatif sur la compréhension.

Lazaraton (2004 : 91) parle de *« teacher gestures »* ou plus généralement de comportement non-verbal enseignant pour désigner ce que nous dénommons kinésique et proxémique. Nous les différencions car d'un côté, la kinésique concerne l'activité musculaire de l'enseignant et d'un autre côté, la proxémique décrit l'utilisation de l'espace par l'enseignant dans sa classe.

Voici résumée l'analyse des comportements paraverbaux et non-verbaux utilisés et utilisables lors de séquences explicatives.

| **Légende** | |
|---|---|
| *En italique* | Technique non observée directement dans notre corpus |

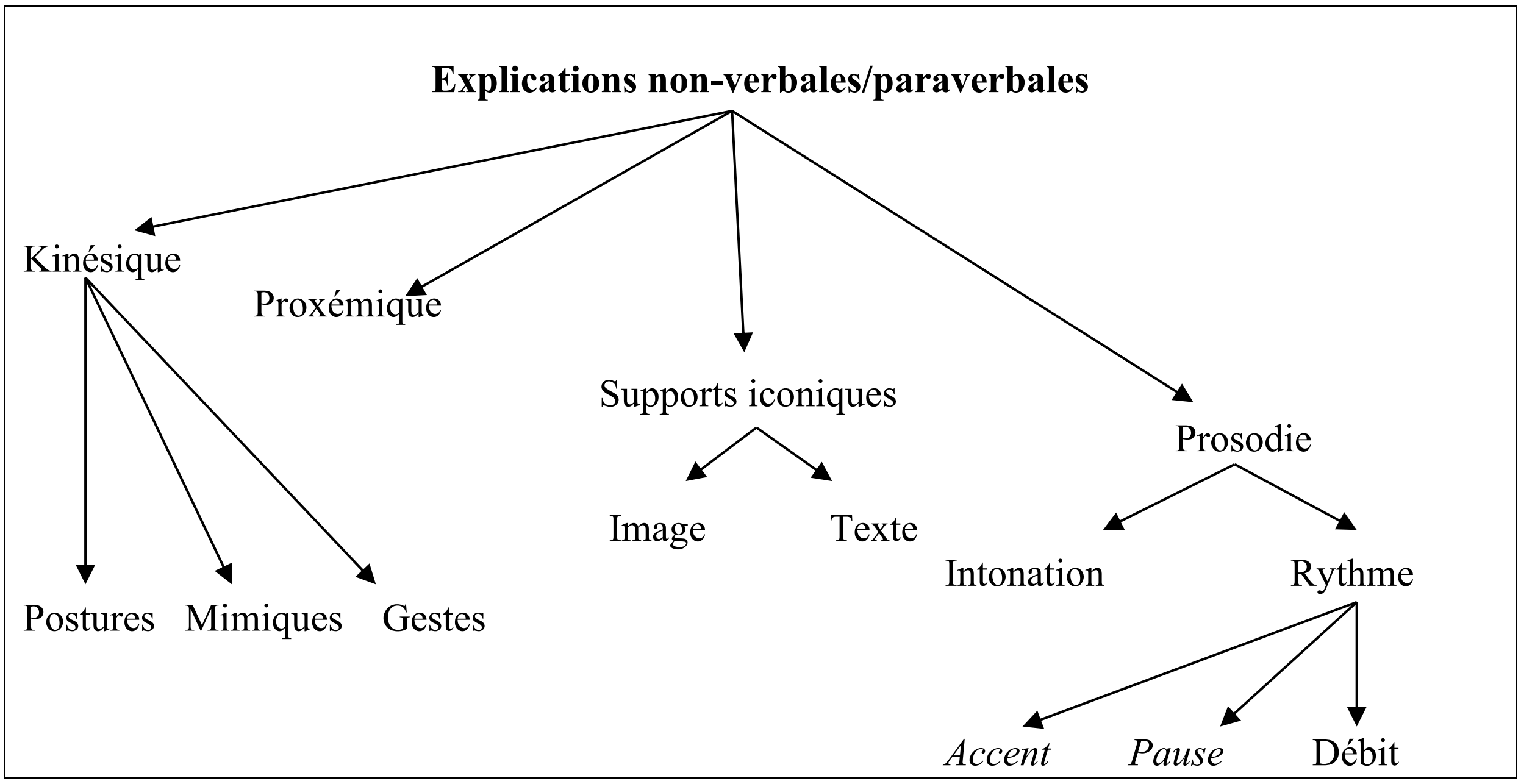

### 6. Traiter automatiquement le discours explicatif en interaction

Notre travail a fait apparaitre des séquences explicatives multimodales qui peuvent être clairement définies : séquence tripartite de compréhension d'un message, utilisant obligatoirement du verbal (langue), facultativement du paraverbal (prosodie) et du non-verbal (posturo-mimo-gestualité). Pour traiter automatiquement ces données multimodales, il conviendra de définir des conventions de transcription qui permettent de retrouver ces différents paramètres et de les associer à une logique sémantique.

**Références :**

# De la complexité de la transcription des interactions multimodales en situations pédagogiques :

## État de l'art des conventions de transcription

Emilie REMOND
TECHNE (EA 6316)

---

Cet état de l'art s'inscrit dans le cadre du projet de recherche TRAIMA (TRaitement Automatique des Interactions Multimodales en Apprentissage). Ce projet de recherche a pour objectif d'explorer le potentiel de traitement automatique des interactions multimodales en situations pédagogiques. Actuellement, l'analyse des interactions verbales, paraverbales et non-verbales est effectuée manuellement par les chercheurs. Très coûteuse en temps, elle pourrait être remplacée par des algorithmes de *Machine Learning* qui les catégoriseraient et les classifieraient.

Une première étape essentielle du travail a été d'établir l'état de l'art des normes de transcription déjà existantes. L'objectif était de s'approprier une convention qui serait adaptée à notre problématique et à nos besoins. Lors de nos recherches, notre attention s'est portée en particulier sur l'approche adoptée par Gaëlle Ferré dans sa thèse[1]. Le travail mené présente effectivement le grand intérêt de tenir compte simultanément des « relations entre discours, intonation et gestualité », tout en déployant une méthodologie concrète. Pour cette raison, ce travail nous a semblé très proche des objectifs que nous cherchions à atteindre. Trois autres conventions de transcription, complémentaires et bien connues des chercheurs, ont également été retenues. Tout d'abord, celle du Groupe Aixois de Recherche en Syntaxe (GARS) fondé à l'Université de Provence dans le milieu des années 1970 avec, à sa tête, Claire Blanche-Benveniste[2] ne traite pas spécifiquement de la gestualité, mais adopte une vision fine des faits d'oralité. Cette approche rejoint celle développée par le centre de recherche belge VALIBEL (Variétés Linguistiques du français en Belgique). Les gestes, sont quant eux, plus communément traités par la convention développée par Lorenza Mondada qui exploite les détails langagiers et corporels[3]. Cet état de l'art sera donc structuré autour de la présentation de ces différentes approches dont nous tenterons de définir les intérêts et les limites.

---

[1] Gaëlle Ferré. Relations entre discours, intonation et gestualité en anglais britannique. Linguistique. Université de la Sorbonne nouvelle - Paris III, 2004. Français. En ligne : https://tel.archives-ouvertes.fr/tel-00135076/document

[2] Sandrine CADDÉO, Frédéric SABIO, Le Groupe Aixois de Recherche en Syntaxe et les recherches actuelles sur le français parlé, Repères DoRiF n.12 - Les z'oraux - Les français parlés entre sons et discours - Coordonné par Enrica Galazzi et Marie-Christine Jamet, DoRiF Università, Roma juillet 2017,
En ligne : http://dorif.it/ezine/ezine_articles.php?id=340

[3] Lorenza Mondada, » Constitution et exploitation de corpus vidéo en linguistique interactionnelle : rendre disponibles les détails multimodaux de l'action située », *Cahiers de praxématique* [En ligne], 54-55 | 2010, document 19, mis en ligne le 01 janvier 2013, consulté le 23 avril 2020. URL : http://journals.openedition.org/praxematique/1183

## 1. La méthodologie développée par Gaëlle Ferré : une approche globale

### 1.1. Principes généraux d'enregistrement, de découpage et transcription

Dans sa thèse, Gaëlle Ferré expose de façon méthodique les différentes étapes qui jalonnent son travail. Il s'agit tout d'abord de justifier le choix des interlocuteurs puis de détailler minutieusement la situation d'enregistrement (en signalant les éventuels biais) avant d'expliciter la consigne. Tout comme la situation, le matériel d'enregistrement, de visionnage et de montage (*Quick Time Player* et *Final Cut Pro*) est également décrit, en explicitant les choix opérés.

A propos des logiciels dédiés à l'étiquetage de la gestualité, la chercheuse en expose deux auxquels elle n'a cependant pas eu accès : TASX et ANVIL. Ces deux outils présentent des fonctionnalités intéressantes, telles que la possibilité de « *visionner la vidéo image par image, d'importer les tires de segmentation effectuées sous Praat et de les compléter par de nouvelles tires d'annotation des gestes. Ceci permet d'aligner le début des gestes sur les tires de texte et de voir immédiatement apparaître les effets de synchronisation ou les décalages entre la gestualité et l'intonation* », (Ferré, 2004 : 26). Si cette précision est impossible à obtenir manuellement, il n'apparait cependant pas essentiel pour la chercheuse de connaitre avec précision la durée d'un décalage entre geste et parole, préférant se concentrer sur ce qui se passe lors des prises de parole ou des silences.

En ce qui concerne l'analyse prosodique, Gaëlle Ferré favorise le logiciel Praat plutôt qu'Anaproz. Si son utilisation apparait simple, les résultats chiffrés de ce dernier manquent parfois de précision, contrairement à Praat. Ce choix opéré, la chercheuse expose ensuite les différentes étapes de découpage du corpus et de transcription : numérisation et étiquetage du signal > extraction et traitement des données > calcul du débit > calcul des plages intonatives > classification des gestes. Ce dernier aspect nous a paru particulièrement intéressant puisque les gestes sont retranscrits en tenant compte de leur signification particulière, raison pour laquelle nous choisissons ici de développer cet aspect.

### 1.2.Penser la classification de gestes signifiants

La question de la description de la gestualité n'est pas récente et peut-être résolue avec plus ou moins de succès. La classification des gestes proposée dès 1987 par Jacques Cosnier[4] propose cependant une distinction fine entre gestes « communicatifs » et gestes « extra-communicatifs ». Certains gestes « communicatifs » apparaissent ainsi comme « quasi-linguistiques » car porteurs de significations liées à des conventions culturelles. Tel est le cas avec des mains jointes et rapprochées du corps signifiant la prière. D'autres gestes « communicatifs » sont dits « syllinguistiques » car ils apparaissent nécessairement associés à la parole. Tel est le cas des « déictiques » qui désignent le référentiel en le pointant du doigt ou encore des « expressifs » lorsque les mimiques du visage accompagnent les propos. Quant aux gestes « extra-communicatifs », ils désignent des gestes de confort, comme le changement de position ou des gestes ludiques lorsque, par exemple, le locuteur manie un objet[5].

Plus récemment, Jean-Marc Colletta (2000) a proposé une classification plus précise de la co-gestualité, en choisissant d'écarter les gestes « extra-communicatifs ». Le chercheur préfère ainsi la terminologie « kinème » à « gestes », par analogie au phonème ou au graphème. Les kinèmes, autonomes ou associés à la parole, sont classés selon leur fonction (discursifs, interactifs, méta-discursifs, syntaxiques ou référentiels) et leur valeur (mimétiques, rythmique, illustratif, anaphorique …). Le rôle du kinème par rapport à la parole est également précisé (substitution, accentuation, illustration, …)[6]. Si cette classification s'avère particulièrement précise et complète pour le relevé des gestes, il n'est cependant pas toujours aisé de distinguer les valeurs des kinèmes, qui peuvent rester sujets à interprétation. Forte de ce constat et consciente de ces limites, Gaëlle Ferré retient cette classification, en y apportant quelques adaptations. Elle choisit ainsi de réintégrer les gestes « extra-communicatifs » de la classification de Jacques Cosnier et d'utiliser le terme de « battement » à la place des gestes rythmiques. Elle conserve les gestes déictiques et anaphoriques, tout comme les notions de mime et de gestes illustratifs.

[4] Cosnier, J., & Kerbrat-Orecchioni, c. (Eds.). (1987). Décrire la conversation. Lyon : Presses Universitaires de Lyon

[5] Pour plus de détails, voir la classification des gestes proposée par J. Cosnier dans Cosnier & Kerbrat Orecchioni (1987 : 296-297).

Finalement, la classification adoptée par Gaëlle Ferré et construite à l'appui des travaux de Jacques Cosnier (1987) et de Jean-Marc Colletta (2000) permet de prendre en compte simultanément le contexte discursif dans lequel le geste est produit (les indices discursifs), les indices prosodiques et leurs valeurs iconiques, et, pour finir, les indices posturo-mimo-gestuels. En ce sens, elle nous parait traduire, tant que faire ce peu, la multitude des indices mis en œuvre conjointement lors des interactions.

## 2. La transcription des gestes par Lorenza Mondada : une convention opérationnelle

Si la méthodologie décrite par Gaëlle Ferré apparait extrêmement rigoureuse et lisible, il nous a semblé difficile de nous approprier sa convention de transcription des gestes, et ce malgré l'intérêt théorique certain qu'elle présente. La réflexion sur les interprétations de gestes parait être trop complexe pour être immédiatement opérationnelle, raison pour laquelle nous nous proposons ici d'en présenter une convention plus usuelle, celle développée par Lorenza Mondada. La linguiste a effectivement explicité ses conventions dans de nombreux articles dans lesquels elles sont utilisées avec clarté (Mondada, 2004, 2005, 2006, 2007, 2008). Nous en présentons ici les grandes lignes[7].

Tout d'abord, les conventions utilisées par Lorenza Mondada visent à représenter les gestes au sens large (actions, mouvements, regards, gestes) simultanément à la parole ou durant des temps de silence. Elles permettent de tenir compte de ce qui se joue dans l'interaction, en examinant non seulement les actions faites par le locuteur mais aussi celles des autres participants à l'échange. Il apparait alors nécessaire d'identifier clairement les personnes produisant les différents gestes à décrire. On distingue ainsi le locuteur en train de produire le geste des autres participants en action, sans qu'ils s'expriment nécessairement. Chaque locuteur est identifié par un symbole spécifique servant également à délimiter le début et la fin des gestes décrits (par exemple, + pour ERE). Lorsque le locuteur est en train de parler, la description du geste n'est pas précédée d'une initiale. A l'inverse, lorsque le geste est produit par un autre participant, ses initiales sont indiquées en minuscule et de façon décalée.

6 Pour plus de détails, voir le tableau « Catégories fonctionnelles de la kinésie communicative », dans la thèse de Gaëlle Ferré, page 43. En ligne : https://tel.archives-ouvertes.fr/tel-00135076/document

7 Pour un aperçu synthétique et complet des normes de transcription élaborées par Lorenza Mondada, voir : http://icar.univ-lyon2.fr/projets/corinte/documents/convention_transcription_multimodale.pdf , consulté le 24 avril 2020.

Les différents participants étant ainsi identifiés en fonction de leur prise de parole, les gestes sont délimités du début à la fin, en tenant compte de leur durée et de leur simultanéité avec les paroles produites. Par exemple, si un geste se prolonge au-delà de la fin de l'extrait, il sera représenté par une double flèche. Puis, il s'agit de décrire ce que Lorenza Mondada appelle la « trajectoire du geste » et qui comprend trois phases : l'amorce, l'apogée et l'éloignement[8]. Pour une vision plus exacte, il est tout à fait possible de renvoyer à une capture d'écran en prenant soin d'indiquer à quel moment précis elle correspond. Ensuite, les lignes doivent être numérotées, étant entendu que la numérotation cible les lignes renvoyant à la prise de parole qui constitue l'étalon temporel sur lequel se calent les gestes. Enfin, des commentaires peuvent être indiqués entre des doubles parenthèses (par exemple, ((en riant))).

Ainsi, la convention proposée par Lorenza Mondada présente l'intérêt de décrire finement les actions produites, sans céder à l'interprétation des gestes. Elle permet également de diviser la description en différentes étapes, ce qui oblige à une certaine rigueur de l'observation.

## 3. Les conventions VALIBEL et GARS : une attention aux marques d'oralité

Dès sa création, le centre de recherche belge VALIBEL (Variétés Linguistiques du français en Belgique) s'est doté de sa propre convention explicite de transcription, selon quatre principes. Tout d'abord, le groupe adopte une orthographe standard, ce qui n'est pas forcément toujours le cas lorsque, par exemple, le transcripteur souhaite mettre en évidence des caractéristiques liées à la prononciation (comme les accents régionaux, par exemple). Cette pratique s'apparente, selon Claire Blanche Benveniste et Colette Jeanjean (1987), à des « trucages orthographiques », dans la mesure où l'orthographe adoptée se veut une représentation de la phonétique des locuteurs. Cependant, cette façon de faire complique la lecture et peut engendrer un effet de stigmatisation lié au non-respect de la norme orthographique (Dister, Simon : 2007). Ensuite, le groupe choisit de ne pas recourir à la ponctuation de l'écrit qui « suggèr[e] une analyse avant de l'avoir faite », (Blanche-Benveniste, Jeanjean, 1987 : 142). Puis, la convention VALIBEL valorise l'oralité des corpus, d'une part en tenant compte des pauses, répétition de mots ou ponctuants (tels que « euh », « bon » …) ou les amorces de morphèmes ; d'autre part, en considérant dans la transcription les tours de parole comme des unités visuelles par défaut. Les chevauchements de parole sont également notés, en les soulignant dans la transcription. Enfin, le groupe VALIBEL s'oriente vers des conventions compatibles avec un traitement informatisé. A ce propos, le groupe VALIBEL opte pour le logiciel Praat pour segmenter et transcrire. Les raisons de ce choix sont multiples et rejoignent celles de Gaëlle Ferré : ce logiciel libre au format ouvert présente effectivement de nombreuses fonctions de segmentation et d'étiquetage.

Pour conclure, les travaux menés par le groupe VALIBEL présentent l'intérêt majeur de ne pas évacuer les marques d'un discours en cours d'élaboration. Ils se rapprochent d'une autre convention traditionnelle, également largement utilisée par les linguistiques : celle de l'équipe du Groupe Aixois de Recherche en Syntaxe (GARS)[9]. Contrairement au groupe VALIBEL, la convention GARS choisit de retenir une orthographe normée en parallèle d'une orthographe phonétique lorsque, par exemple, les locuteurs ne sont pas francophones. Tout comme le groupe VALIBEL, les marques spécifiques de l'oralité sont notées (« euh », « ben » …) et certaines indications (telles que les rires) peuvent être indiqués sous forme de notes. Le signe « X » permet quant à lui d'indiquer les passages inaudibles. De même, les chevauchements sont indiqués par des soulignements. Fidèle à l'idée selon laquelle la ponctuation présage l'interprétation du transcripteur, la convention GARS évite aussi d'inscrire les marques de ponctuation.

[8] La phase d'amorce est représentée par le symbole ….. et la phase d'éloignement par ,,,,,, . L'apogée représente le geste au moment où il est pleinement déployé et est alors directement nommé (par exemple, « pointe »).

[9] Voir en particulier les travaux de Claire Blanche-Benveniste : Blanche-Benveniste, C. (1997). Approches de la langue parlée en français, Paris/Gap, Ophrys.

**Conclusion :**

Au terme de cet état de l'art, plusieurs remarques s'imposent. Tout d'abord, en ce qui concerne le verbal, les marques d'oralité sont essentielles car la transcription d'un oral ne saurait être une traduction vers l'écrit. En ce sens, nous rejoignons la position adoptée par les conventions GARS et VALIBEL. Pour cette transcription du verbal, un consensus semble établi pour l'usage de Praat, logiciel libre présentant de nombreuses fonctionnalités pratiques. Ensuite, la transcription des gestes, qu'ils apparaissent simultanément à la prise de parole ou indépendamment d'elle, sont des marques essentielles de l'interaction révélatrices de ce qui se joue dans les échanges. La difficulté semble alors de décrire sans interpréter, d'où le choix de convention adoptée par Lorenza Mondada. Enfin, la méthodologie déployée par Gaëlle Ferré en cinq étapes distinctes, de la numérisation à la classification des gestes, permet de prendre en compte l'ensemble des relations entre discours, intonation et gestualité. Chacune de ces conventions présente ainsi un intérêt certain. Cependant, il semble que pour être réellement opérationnelle, une convention doit être éprouvée au terme d'un tâtonnement propre à une démarche expérimentale et au projet d'une équipe. Reste ainsi à mettre à l'épreuve ces différentes conventions afin de se les approprier en pratique.

**Références :**

# Une séquence explicative à l'épreuve des conventions de transcription :
## Traitement manuel et variabilité du traitement

Julie RANCON
FORELLIS (EA 3816)
Emilie Remond
TECHNE (EA 6316)

L'analyse de deux transcriptions manuelles d'une même séquence pédagogique – réalisées respectivement par Julie Rançon et Émilie Remond – permet de mettre en lumière la diversité des pratiques de transcription dans le domaine de l'analyse multimodale. À partir d'un extrait d'interaction en classe portant sur la notion de *résidence secondaire*, les deux versions proposent des représentations fidèles mais différentes de la scène, révélant des choix techniques, théoriques et interprétatifs distincts. Cette comparaison permet de réfléchir aux enjeux méthodologiques liés à la transcription manuelle et à sa fonction centrale dans le traitement des données interactionnelles.

**Séquence explicative :**

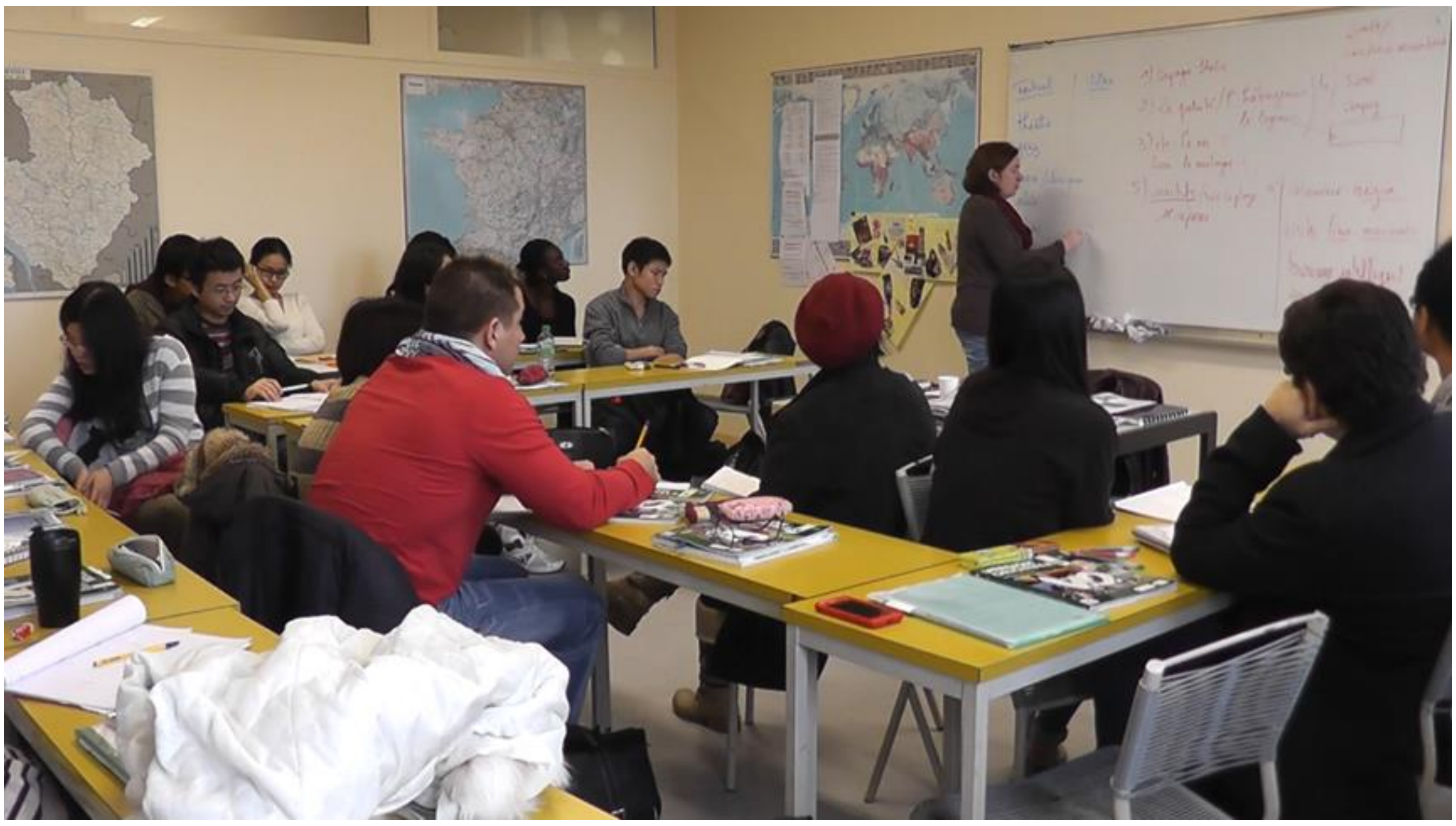

L'extrait analysé provient d'une séance de FLE enregistrée dans une classe de niveau B1. L'enseignante interroge les apprenants sur la signification du syntagme *résidence secondaire*. La séquence dure environ 27 secondes et présente les caractéristiques d'une explication tripartite :

- Ouverture problématisante : interrogation adressée à la classe ;
- Noyau explicatif : définition co-construite avec une apprenante, soutenue par des gestes iconiques et déictiques ;
- Clôture évaluative : validation et élargissement à une généralisation.

Nous avons transcrit manuellement cette même séquence selon trois conventions :

**a. ICOR**

- Segmentation en tours de parole.
- Intégration de gestes co-occurrentiels.

- Annotation synchronisée (verbal/paraverbal/non-verbal).
- Utilisation de chevrons, crochets, indications temporelles.

**b. Mondada**

- Encodage rigoureux de la trajectoire gestuelle (préparation, apogée, retrait).
- Identification des locuteurs par symboles spécifiques.
- Temporalité très fine mais nécessite un alignement vidéo systématique.
- Peu d'interprétation, très descriptif.

**c. Tellier**

- Catégorisation fonctionnelle des gestes : déictiques, illustratifs, rythmiques.
- Intégration des dimensions prosodiques (pauses, intonation, débit).
- Vision fonctionnelle et interactionnelle du geste.
- Meilleure lecture didactique, mais plus interprétative.

## Transcription manuelle Julie Rançon

**TRANSCRIPTION MANUELLE ICOR**

**maison secondaire**

((00:03))

ENS <donc il vous manque ça ((sur le tableau, fait un trait sous une liste de mots))>

ENS <((ENS pointe du stylo le tableau où est noté 'résidence secondaire'. Elle se retourne vers la classe, croise les bras tout en portant la main droite fermée tenant le stylo, sur sa bouche. Elle énonce la phrase suivante en même temps qu'elle fait ces gestes)) qu'est-c'que ça veut dire résidence secondaire/>

(2.3)

ENS [qu'est-c'que ça veut dire/ ]

XXF [(inaud.) ]

ENS [<(( désigne XXF du doigt, geste de pointage de la main droite))>]=

XXF =<((geste déictique de ENS))c'est la deuxième maison coté euh> à côté de la mer la plage (inaud.)(0.5)

XXF =c'est la deuxième maison coté euh> à côté de la mer la plage (inaud.)(0.5)

ENS ouais c'est la <deuxième maison ((stylo dans la main, ENS montre l'index et le majeur, deux fois))> (0.3) <beaucoup de gens sont propriétaires ((main droite tenant toujours le stylo vers la gauche))> de <deux ((stylo dans la main, ENS montre l'index et le majeur, deux fois))> maisons (.) la <((geste de la main droite qui va vers la gauche, bras toujours croisés)) maison principale> (0.3) et <((geste de la main droite qui va vers la droite, bras toujours croisés)) la maison de vacances> <((geste de la main droite allant de la gauche vers la droite)) la maison secondaire> <((geste de la main droite montrant l'index et le majeur)) la deuxième> (0.4) <((geste de la main droite allant vers l'avant)) pour les vacances> (0.7) <((geste de la main droite vers les élèves de droite)) et ça peut être n'importe où hein> <((geste faisant le signe de l'infini)) à la mer à la montagne à la campagne> <((geste de la main droite allant vers la droite, loin)) dans une au:tre ville> ça peut être aussi <((geste de pince ouverte avec le pouce et de l'index)) un:> <((gestes de recherche lexicale)) appartement dans une grande ville> (0.3) <((geste de la main gauche, désigne FAT, acquiesce de la tête)) pour les gens qui> <((les deux mains sont rapprochées vers la poitrine, légère gesticulation des doigts)) aiment faire comme fatia du shopping <((acquiescement de la tête)) (0.5)> <((les deux mains vont vers l'avant)) si vous aimez plutôt la tranquillité> <((les deux mains vont vers la gauche)) vous allez plutôt dans le> <((rassemblement des mains en poing))(0.2) sud de la vienne> <((geste d'acquiescement de la tête)) au bord de l'eau:> (0.6) maison secondaire((fin du geste en poing des mains))>(0.2) <((geste de pointage du tableau par la main gauche)) donc faudra chercher ça ici hein> (0.6) <((fait un rectangle au tableau à l'aide du premier trait tracé))quel est le dernier type> (0.5) d'hébergement <((trace au tableau une flèche qui va de 'le logement' au rectangle vide dessiné)) (.)> ((se retourne vers les élèves)) qui est cité dans le document>

((00 :50))

## TRANSCRIPTION ICOR + typologie des gestes de TELLIER

## maison secondaire

((00:03))

ENS <donc il vous manque ça ((sur le tableau, fait un trait sous une liste de mots))>

ENS <((geste de pointage sur 'résidence secondaire' au tableau. Elle se retourne vers la classe, croise les bras tout en portant la main droite fermée tenant le stylo, sur sa bouche. Elle énonce la phrase suivante en même temps qu'elle fait ces gestes)) qu'est-c'que ça veut dire résidence secondaire/>

(2.3)

ENS [qu'est-c'que ça veut dire/ ]

XXF [(inaud.) ]

ENS [<(( geste de pointage vers XXF))>]=

XXF =<(( geste déictique de ENS))c'est la deuxième maison coté euh> à côté de la mer la plage (inaud.)(0.5)

ENS ouais c'est la <deuxième maison ((stylo dans la main, ENS montre l'index et le majeur, deux fois. Geste iconique))> (0.3) <beaucoup de gens sont propriétaires ((main droite tenant toujours le stylo vers la gauche. Geste métaphorique))> de <deux ((stylo dans la main, ENS montre l'index et le majeur, deux fois. Geste iconique))> maisons (.) la <((geste métaphorique de la main droite qui va vers la gauche, bras toujours croisés)) maison principale> (0.3) et <((geste métaphorique de la main droite qui va vers la droite, bras toujours croisés)) la maison de vacances> <((geste métaphorique de la main droite allant de la gauche vers la droite)) la maison secondaire>

<((geste iconique de la main droite montrant l'index et le majeur)) la deuxième> (0.4) <((geste métaphorique de la main droite allant vers l'avant)) pour les vacances> (0.7) <((geste déictique ou interactif de la main droite vers les élèves de droite)) et ça peut être n'importe où hein> <((geste métaphorique ou de battement faisant le signe de l'infini)) à la mer à la montagne à la campagne> <((geste métaphorique de la main droite allant vers la droite, loin)) dans une au:tre ville> ça peut être aussi <((geste iconique de pince ouverte avec le pouce et de l'index)) un:> <((gestes butterworth de recherche lexicale)) appartement dans une grande ville> (0.3) <((geste de pointage de la main gauche, désigne FAT, acquiesce de la tête)) pour les gens qui> <((geste de battement, les deux mains sont rapprochées vers la poitrine, légère gesticulation des doigts)) aiment faire comme fatia du shopping <((acquiescement de la tête, geste emblème, conventionnel)) (0.5)> <((geste métaphorique, les deux mains vont vers l'avant)) si vous aimez plutôt la tranquillité> <((geste métaphorique, les deux mains vont vers la gauche)) vous allez plutôt dans le> <((rassemblement des mains en poing))(0.2) sud de la vienne> <((geste conventionnel d'acquiescement de la tête)) au bord de l'eau:> (0.6) maison secondaire((fin du geste en poing des mains))>(0.2) <((geste de pointage du tableau par la main gauche)) donc faudra chercher ça ici hein> (0.6) <((fait un rectangle au tableau à l'aide du premier trait tracé))quel est le dernier type> (0.5) d'hébergement <((trace au tableau une flèche qui va de 'le logement' au rectangle vide dessiné)) (.)> ((se retourne vers les élèves)) qui est cité dans le document>

((00 :50))

## TRANSCRIPTION ICOR + typologie Mondada

## maison secondaire

((00:03))

ENS <donc il vous manque CA ((sur le tableau, fait un trait sous une liste de mots))>

ENS +qu'est-c'que ça veut dire résidence secondaire/(2.3)+
+pointe sur 'résidence secondaire' au tableau. ENS se retourne vers la classe, croise les bras tout en portant la main droite fermée tenant le stylo, sur sa bouche (position 1) -------------------------+

ENS [qu'est-c'que ça veut dire/]

XXF [(inaud.) ]+(inaud.) c'est la deuxième+
EN +pointe-----------------------+

XXF +maison coté euh+ +à côté de la mer la plage (inaud.)(0.5)+
EN +déictique-----,,,+ +position 1-----------------------------------+

ENS ouais c'est la +deuxième maison+(0.3)
+iconique --------+

ENS +beaucoup de gens sont propriétaires+ de +deux + maisons (.)
+métaphorique---------------------------+ +iconique+

la +maison principale+ (0.3) et +la maison de vacances+
+ métaphorique------+ +métaphorique-----------+

+la maison secondaire+ +la deuxième+(0.4) +pour les vacances+(0.7)

+métaphorique----------+ +iconique-----+ +métaphorique-------+

+et ça peut être n'importe où hein+ +à la mer à la montagne à la
+déictique ou interactif--------------+ + métaphorique ou battement---->

campagne+ +dans une au:tre ville+ ça peut être aussi +un: +
---------+ +métaphorique------------+ +iconique+

+appartement dans une grande ville+(0.3) +pour les gens qui+
+ butterworth------------------------+ +pointage-------,,,,+

+aiment faire comme fatia du shopping+ +(0.5) +
+battement------------------------------+ +emblème+

+si vous aimez plutôt la tranquillité+ + vous allez plutôt dans le+
+métaphorique---------------------------+ +métaphorique----------------+

sud de la vienne (0.2) +au bord de l'eau:+ (0.6) maison secondaire
+conventionnel------+

(0.2) donc +faudra chercher ça ici hein+(0.6) <quel est le dernier
+pointage---------------------+

type ((fait un rectangle au tableau à l'aide du premier trait tracé))> +(0.5) d'hébergement <(0.2)((trace au tableau une flèche qui va de 'le logement' au rectangle vide dessiné))>

+qui est cité dans le document+
+se retourne vers les élèves-----+

((00 :50))

## Transcription manuelle Emilie Remond

### Transcription ICOR et MONDADA

ENS : *

ENS : *donc il vous manque CA/* Qu'est-ce que ça veut dire résidence secondaire/ (0.4)
*écrit au tableau * * se retourne vers les étudiants et ferme les bras*

Qu'est-ce que [ça veut &

ETU 1 : [ C'est le &&
ENS : &* dire* ]
*pointe*

ETU1 : c'est le deuxième mai[son &
ens *hoche la tête*

ETU2 : [&& zon]

ETU1 : euh: côté de euh à côté de la mer [ (inaud.)]

ENS : *c'est la DEUXIÈME maison * Beaucoup de gens sont propriétaires de *DEUX*
*pointe deux doigts vers le haut * *lève deux doigts vers le haut*

maisons \(.) *la maison principale* et *la maison de vacances* (.)\ la maison secondaire
*main vers le bas* * main vers le haut*

(.) *la deuxième*(.) *POUR les vacances* \ (.) *et ça peut |être n'importe où hein*/ *A la
*lève deux doigts* *ouvre la main* * pointe avec la main* * vague

mer, à la montagne à la campagne dans une autre vi:lle *ça peut aussi être un:
avec la main----------------------------------------------------------*

appartement dans une *GRANDE* ville (.) *pour les gens qui aiment faire *comme Fatia
*Ouvre les main* * pointe------------------------------*

du shopping(.) si vous aimez plutôt la tranquillité *vous allez dans le*: sud de la Vienne
*Mains vers la gauche*

au bord de l'eau: \ (.) * Maison secondaire* (.)
*geste clôture*

*DONC faudra chercher* ça *ici* hein / ((écrit au tableau)) quel est (.)le dernier type (.)
*pointe au tableau* *écrit------------------------------------------------------------------

d'hébergement (.) *qui est cité (.) dans le document* /
---------------------- *ferme mains se retourne*

## Comparer deux transcriptions manuelles d'une même séquence : différences, convergences et enseignements méthodologiques

Les deux transcriptions partagent une base verbale commune. Elles s'appuient sur le même segment de discours, issu d'une séance explicative centrée sur la notion de « résidence secondaire », avec un enchaînement verbal quasi identique. Dans les deux cas, les composantes non verbales de l'interaction sont prises en compte. Les gestes, les postures, les orientations du corps et certains éléments prosodiques sont décrits, montrant un souci partagé de capter la dimension multimodale de la scène. Chaque transcription intègre un repère temporel ((00:03) à ((00:50)), situant l'extrait dans un enregistrement plus large. Cela permet d'envisager l'extrait comme une séquence située dans un déroulé global. Mais des différences structurelles et analytiques apparaissent rapidement.

La transcription d'Émilie Remond adopte un format linéaire, typé ICOR, dans lequel les gestes sont encadrés par des astérisques (*). Le texte est organisé de manière verticale, avec peu de chevauchements. Les gestes sont décrits de façon référentielle : « ouvre la main », « pointe », « lève deux doigts vers le haut ». Il s'agit donc d'une transcription descriptive qui vise à représenter la scène en s'appuyant sur ce qui est visuellement observable.

La transcription de Julie Rançon, quant à elle, mobilise une structuration davantage typée Mondada. Elle recourt à des codes synchroniques (+) qui permettent d'aligner les gestes à l'énoncé verbal. Cette organisation favorise une lecture simultanée des différentes modalités, en soulignant les chevauchements, les enchaînements et les synchronies. Surtout, cette version intègre une typologie gestuelle explicite. Les gestes sont catégorisés selon leur fonction sémiotique, avec des étiquettes telles que « déictique », « iconique », « métaphorique », « emblème », « butterworth », ou encore « battement ». On passe donc d'une simple description du visible à une lecture interprétative des gestes, fondée sur les travaux de Tellier et Mondada.

Cette différence de granularité descriptive a des effets concrets sur l'analyse. Par exemple, lorsqu'il s'agit de désigner le mot au tableau, Émilie Remond note :
*« pointe »*

Tandis que Julie écrit :
+» pointe sur 'résidence secondaire' au tableau »+
+déictique+

La deuxième transcription rend explicite la fonction sémiotique du geste (déictique) et précise son ancrage spatial, ce que la première laisse implicite.

Autre exemple lors de la reformulation avec le syntagme « deux maisons », Émilie Remond écrit :
*« pointe deux doigts vers le haut »*
*« lève deux doigts vers le haut »*

Julie l'interprète comme :
+iconique+

Elle traduit ainsi le geste comme une représentation iconique de la quantité évoquée, c'est-à-dire comme un geste mimant le contenu sémantique du mot, et non comme une simple illustration manuelle.

Dans la séquence « comme Fatia, pour faire du shopping », Émilie note :
*« ouvre les mains »*
*« pointe »*

Julie propose :
+pointage+
+battement+
+emblème+

Ce triple codage permet de saisir la densité pragmatique du geste : désignation de l'apprenant, mouvement accentuant le propos, et usage d'un geste conventionnel compris dans le cadre socio-discursif.

L'interprétation des gestes dépend donc de la posture analytique du transcripteur. Chaque transcription traduit des choix, qui dépendent de la formation du chercheur, de ses cadres théoriques, et de l'objectif poursuivi. La posture du chercheur-transcripteur influe directement sur ce qui est relevé, ignoré, catégorisé ou simplement décrit. La transcription d'Émilie Remond est particulièrement adaptée à une lecture pédagogique ou illustrative. Elle documente les gestes de l'enseignant de manière observable, dans une logique de description référentielle. Celle de Julie Rançon permet une analyse interactionnelle plus fine, orientée vers les dimensions sémiotiques et pragmatiques de l'agir enseignant. Elle est plus adaptée à des finalités d'analyse multimodale approfondie. La typologisation des gestes, telle qu'elle apparaît dans la transcription de Julie Rançon, permet une comparabilité scientifique. La présence de catégories comme « métaphorique », « iconique », « déictique » ou « butterworth » rend possible la mise en correspondance de segments similaires à travers plusieurs corpus, langues ou contextes pédagogiques. Cela favorise la systématisation des observations, la constitution de typologies stabilisées, et l'interopérabilité des analyses. La structuration synchronique de type Mondada permet en outre d'augmenter la lisibilité analytique. En alignant les gestes, la parole et les déplacements sur un même plan temporel, elle facilite l'identification des ajustements multimodaux, des effets d'amplification ou de reformulation, et des dynamiques d'alignement entre enseignant et apprenant. Ce type de transcription rend visible la co-énonciation incarnée qui est au cœur de l'interaction pédagogique.

Enfin, cette comparaison révèle une variabilité intersubjective inévitable mais féconde. Deux chercheurs transcrivant une même séquence peuvent produire des versions différentes. Cette variabilité ne reflète pas un manque de rigueur, mais la richesse des lectures possibles. À condition d'être assumée, motivée et contextualisée, elle constitue un levier d'analyse et non un défaut méthodologique. La transcription manuelle d'interactions multimodales apparaît ainsi comme une activité technique, interprétative et théorique. Elle ne consiste pas à reproduire une séquence, mais à modéliser une situation d'interaction, en en fixant certaines dimensions jugées saillantes, signifiantes ou analysables. Il ne s'agit donc pas de chercher une transcription idéale, mais de rendre compte de la logique de transcription adoptée, en la mettant en cohérence avec la posture de recherche, les objectifs visés, et le cadre théorique mobilisé. Plutôt que d'uniformiser les pratiques, il semble plus pertinent de former les chercheurs à la réflexivité transcriptionnelle, d'expliciter les choix de codage dans les publications, de favoriser des protocoles partagés dans les équipes sans en faire des normes rigides, et d'envisager la transcription comme une étape de construction des données, et non comme une opération neutre. C'est à cette condition que la transcription manuelle pourra assumer pleinement son rôle épistémologique dans la recherche en analyse multimodale.

**Caractéristiques relevées dans les deux transcriptions**

| Critère | Transcription Émilie | Transcription Julie |
|---|---|---|
| **Format de transcription** | Linéaire, typée ICOR avec gestes entre astérisques * | Structuration Mondada avec codes +, commentaires synchronisés aux tours de parole |
| **Organisation spatiale** | Texte organisé de manière verticale, peu de chevauchements | Présentation en colonne synchronisée, permettant de voir simultanéité des gestes et de la parole |
| **Granularité descriptive des gestes** | Gestes décrits de façon descriptive : « ouvre la main », « pointe » | Gestes analysés typologiquement : iconique, métaphorique, déictique, emblème, butterworth, battement… |
| **Typologie gestuelle** | Absente – les gestes sont référentiels | Présente – s'inspire clairement de Tellier et Mondada pour catégoriser les gestes |
| **Interprétation sémiotique** | Limitation à la description observable | Analyse fonctionnelle intégrée (geste de reformulation, de focalisation, d'interaction…) |
| **Synchronisation verbalo-gestuelle** | Faible, parfois implicite | Très structurée : les gestes sont alignés à l'énoncé verbal par codes synchroniques + |

| Critère | Transcription Émilie | Transcription Julie |
|---|---|---|
| **Effet de clôture ou mise en scène** | Mention d'un « geste de clôture » final, peu développé | Plus d'attention à la dynamique finale : séquences de pointage, recadrage, mouvement vers le tableau et retour vers élèves |

## Discussions sur la transcription manuelle des corpus

Les transcriptions d'Émilie Remond et Julie Rançon illustrent de manière exemplaire la diversité des approches possibles dans le traitement manuel des données multimodales. Ces variations ne tiennent pas seulement aux outils ou aux conventions choisies, mais relèvent aussi de la posture du chercheur, de sa sensibilité théorique et des objectifs analytiques poursuivis.

La première différence notable réside dans la granularité descriptive adoptée. Une transcription de type ICOR, comme celle proposée par Émilie Remond, s'appuie sur une description relativement neutre des gestes, en recourant à des formulations référentielles telles que « ouvre les mains » ou « lève deux doigts ». À l'inverse, la transcription enrichie selon une typologie gestuelle, comme celle de Julie Rançon, propose une lecture sémiotique plus fine. Le geste devient alors un signe analysé pour sa fonction : déictique, iconique, métaphorique, emblématique, voire relevant des gestes dits de Butterworth (liés à l'hésitation ou à la recherche lexicale). Ainsi, là où Émilie Remond écrit simplement *« lève deux doigts vers le haut »*, Julie Rançon code ce même geste comme *iconique*, le lisant comme une représentation visuelle de la quantité « deux » évoquée dans l'énoncé. Ce passage de la description à l'interprétation engage une inférence théorique, qui repose sur la formation du transcripteur, sa familiarité avec les systèmes gestuels et son positionnement analytique.

La structure temporelle de la transcription constitue un autre point de divergence. La transcription linéaire d'Émilie Remond suit une temporalité approximative, structurée autour des tours de parole et ponctuée de notations parenthétiques. En revanche, la transcription de Julie Rançon explicite systématiquement les synchronies entre parole, gestes et postures à l'aide de notations synchroniques de type Mondada. Chaque mouvement gestuel est aligné à un segment verbal, ce qui permet de visualiser clairement les chevauchements, les correspondances et les ruptures temporelles. Par exemple, l'alignement du geste +*pointe sur 'résidence secondaire'*+ avec la question +*qu'est-ce que ça veut dire résidence secondaire*+ rend manifeste la coordination entre la modalité verbale et la modalité gestuelle, soulignant leur rôle conjoint dans la co-construction du sens. Ce type de structuration suppose une granularité temporelle fine, souvent centésimale, généralement facilitée par l'usage de logiciels comme ELAN ou Anvil.

Le rôle attribué aux gestes dans l'analyse varie également selon les approches. Certains les considèrent comme de simples compléments illustratifs à la parole, d'autres comme des entités sémiotiques à part entière, parfois indépendantes du verbal. Dans la transcription d'Émilie Remond, les gestes apparaissent essentiellement comme des appuis visuels ponctuels, peu catégorisés fonctionnellement. En revanche, dans celle de Julie Rançon, ils sont interprétés comme des actes interactionnels : pointage assignatif vers une apprenante (+*déictique*+), geste de clôture d'explication (+*conventionnel*+), geste de battement marquant une emphase ou un engagement discursif (+*battement*+), etc. Ces lectures fonctionnelles permettent de mieux comprendre comment l'enseignant mobilise des ressources corporelles pour guider l'attention, structurer le discours ou gérer la dynamique de classe.

Cette variabilité traduit également la subjectivité inhérente à toute transcription. Deux chercheurs observant une même séquence ne produiront pas nécessairement les mêmes descriptions, ne détecteront pas les mêmes gestes, et n'en proposeront pas les mêmes interprétations. Cette variabilité intersubjective s'explique par de nombreux facteurs. Le bagage disciplinaire du transcripteur (linguiste, didacticien, interactionniste…), sa connaissance du contexte observé (sa propre classe ou celle d'un collègue), ses objectifs méthodologiques (exhaustivité ou focalisation), ou encore les théories convoquées (sémiotique gestuelle, cognition incarnée, analyse conversationnelle…). Transcrire, c'est inévitablement projeter un cadre de lecture sur les données : la transcription devient alors un acte de modélisation, qui construit l'objet d'étude autant qu'il le restitue.

Enfin, il convient de rappeler que la nature même de la transcription dépend des visées analytiques poursuivies. Une transcription brute, référentielle, peut suffire à un travail de documentation ou d'archivage. Une transcription enrichie par une typologie gestuelle permettra une analyse sémiotique fine. Une transcription synchronique et multicouche, comme celle proposée par Julie, sera nécessaire pour étudier la co-construction multimodale du sens ou les ajustements interactionnels. Une autre, centrée sur la posture et les supports matériels, conviendra mieux à une étude didactique de la médiation. Dans tous les cas, la transcription n'est pas une opération technique neutre. Elle constitue une première étape interprétative, qui conditionne la lecture des données et les résultats qui pourront en être tirés.

La transcription manuelle des corpus multimodaux repose ainsi sur un équilibre délicat entre conventions collectives, choix méthodologiques et subjectivité individuelle. Elle représente un espace d'interprétation, voire de recontextualisation, où le chercheur engage ses représentations, ses cadres d'analyse et ses attentes. Loin de devoir être standardisée, la variabilité transcriptionnelle peut être reconnue comme un indice de réflexivité méthodologique, à condition qu'elle soit explicitée, justifiée et mise en cohérence avec les finalités de la recherche. À travers l'examen de ces deux transcriptions, on comprend mieux que ce n'est pas tant ce qui est transcrit que *comment* et *pourquoi* cela l'est, qui fonde la valeur heuristique d'une transcription.

| | ANVIL | CLAN | ELAN | EXMARALDA | PRAAT | TRANSANA |
|---|---|---|---|---|---|---|
| **PageWeb** | http://www.dfki.uni-sb.de/~kipp/anvil/ | http://childes.psy.cmu.edu/ | http://www.lat-mpi.eu/tools/elan | http://exmaralda.org/ | http://www.praat.org | http://www.transana.org/ |
| **Auteurs** | Michael Kipp (Université de la Sarre), DFKI (Deutsches Forschungszentrum für Künstliche Intelligenz) | Brian MacWhinney et Leonid Spektor, Carnegie Mellow University | Birgit Hellwig (auteur original), Dieter Van Uytvanck, Micha Hulsbosch (auteurs des mises à jour) | Thomas Schmidt & Kai Wörner (Universität Hamburg) | Paul Boersma et David Weenink, Institute of Phonetic Sciences, University of Amsterdam | Chris Fassnacht (auteur original), David K. Woods – University of Wisconsin-Madison (aujourd'hui). |
| **Plateforme** | Windows, Macintosh, Unix | Macintosh, Windows, Unix | Macintosh, Windows, Linux | Macintosh, Windows, Linux | Macintosh, Windows, Linux | Macintosh, Windows |
| **Disponibilité** | Demande de téléchargement à l'auteur par e-mail | Freeware | Freeware | Téléchargement avec un mot de passe fourni par l'auteur | Freeware | Payant |
| **Format d'entrée** | Audio : wav | Audio: aif, aiff, wav, mp3 | Audio : wav | Audio : wav | Audio : wav, mp3 | Audio : mp3, wav, wmv, mov |
| | Vidéo : avi, mov | Vidéo : mpeg, mpg, dat, mov | Video : mpeg1-2, mov | Video : mpeg | Vidéo : Non | Video : mpeg1-2, avi, mov… |
| **Format natif** | Anvil (.anvil) | CHAT (.cha) & CA (.ca) | Elan (.eaf) | XML (.xml) avec trois DTD associées | Textgrid (.textgrid) | XML (base de données) avec sa DTD, RTF (transcription) |
| **Player intégré** | Quicktime | Quicktime | | | | Windows Media Player, Quicktime |
| **Import** | Praat (Textgrid, PitchTier) | elan>chat, praat>chat | shoebox>elan, chat>elan, transcriber>elan | TASX, Praat, HIAT-DOS | NON | RTF |
| **Export** | Fichier texte | chat>ca, chat>elan, chat>praat, chat>xmar | elan>shoebox, elan>chat, elan>transcript text, … | Praat, Elan, TASX Annotator | NON | XML (base de données) |
| **Edition** | NON | Impression du texte seul sans time code audio ou video | Création d'un fichier excel avec les time code et le texte | RTF, HTML, PDF | pdf | PDF, RTF |
| **Avantage** | Hiérarchie dans les catégories | Possibilité de traiter des fichiers vidéo de grande taille | Possibilité de lire deux flux d'images vidéo en même temps, et même de les synchroniser | Possibilité de créer des fichiers html, pdf et rtf avec l'alignement entre le temps et les annotations | Peut traiter des fichiers audio de plusieurs heures | Possibilité de fixer les erreurs à partir des fichiers XML |
| | Import de transcriptions Praat | Visualisation en temps réel de l'alignement transcription/vidéo | Annotation et codage possible grâce à des lignes de tiers supplémentaire | DTD apparemment assez génériques pour être utilisées selon nos besoins. | Logiciel largement utilisé | 2 versions : MU et SU |
| | Annotation et codage possible grâce à des lignes de tiers supplémentaire | Impossibilité de lire deux flux d'images vidéo en même temps, ni de les synchroniser | Hiérarchisation possible des tiers (avec notion de tiers « dépendantes ») | | API utilisable à l'intérieur des tiers et affichage à l'écran | Gestion d'un nombre illimité de sources vidéo/audio |
| | Recherches statistiques par une syntaxe de commande | | | | | Visionnage d'une suite de clips possible |
| | s particulière | | | | | |
| | Annotation et codage possible grâce a des tiers intermédiaires (précédé par signe %) | | | | | |
| | Grande communauté d'utilisateurs | | | | | |
| **Inconvénient** | Pas de format MPEG1 en entrée | | | Pas d'import de fichiers CLAN ou Anvil | Non transcription de fichiers vidéo. | Nécessite un système d'indexation pour les clips |
| | Pas de vidéo de plus 10min | | | | | |
| | Arrêt fréquent de l'application | | | | | |
| | | | | | | |
| **Source : http://www.icar.cnrs.fr/sites/corinte/** | | | | | | |

Le tableau présenté synthétise les caractéristiques techniques, fonctionnelles et ergonomiques de six logiciels majeurs d'annotation utilisés dans le traitement des corpus multimodaux : **ANVIL, CLAN, ELAN, EXMARaLDA, PRAAT** et **TRANSANA**. Il permet de mettre en évidence la complémentarité mais aussi les limites de ces outils en fonction des objectifs d'analyse (linguistique, interactionnelle, prosodique, didactique…) et des types de données (audio, vidéo, transcription textuelle, alignement temporel).

**Compatibilité des formats et accessibilité**

Tous les outils étudiés sont disponibles sur les principales plateformes (Windows, Mac, Linux), à l'exception d'ANVIL qui nécessite un contact direct avec l'auteur pour l'obtention du logiciel. La majorité sont gratuits (freeware), sauf TRANSANA qui demeure un logiciel payant, ce qui peut constituer un frein pour certains laboratoires ou chercheurs indépendants.
En ce qui concerne les formats audio/vidéo, on observe une prise en charge étendue des formats .wav et .mp3 pour l'audio. Pour la vidéo, ANVIL, ELAN et TRANSANA se démarquent par leur capacité à gérer des formats variés (.avi, .mov, .mpeg…), alors que PRAAT, centré sur le traitement acoustique, ne permet pas de travailler sur la vidéo, ce qui limite son usage dans l'analyse des gestes co-verbaux et des interactions filmées.

## Fonctionnalités d'annotation et de synchronisation

En matière d'annotation multimodale, ELAN et EXMARaLDA apparaissent comme les outils les plus complets et flexibles. Ils permettent d'annoter sur plusieurs tiers hiérarchisés, avec possibilité de créer des dépendances, d'ajouter des métadonnées ou encore de gérer plusieurs pistes temporelles en parallèle. Ces fonctionnalités sont essentielles pour l'analyse des séquences explicatives, qui mobilisent simultanément la parole, le geste, le regard, et parfois des supports visuels ou écrits. ELAN offre en outre la possibilité de lire deux flux vidéo simultanément, fonctionnalité particulièrement utile pour comparer les mouvements corporels d'un enseignant et les réactions d'un apprenant dans une situation pédagogique. TRANSANA, de son côté, permet de gérer un nombre illimité de sources multimédia et offre une fonction de visionnage de clips successifs, avantageuse pour l'étude de corpus segmentés en épisodes. CLAN, bien que moins ergonomique, se distingue par sa capacité à produire des recherches statistiques avancées via une syntaxe de commande dédiée. Cet atout est précieux dans les analyses quantitatives de grands corpus (notamment issus de CHILDES ou d'autres corpus conversationnels).

## Interopérabilité et exportation

L'interopérabilité des formats est un critère central dans les pratiques de recherche actuelles. ELAN se démarque nettement par sa capacité à importer et exporter dans de nombreux formats (Shoebox, Praat, CHAT, etc.), facilitant l'intégration avec d'autres outils ou le travail collaboratif entre chercheurs. EXMARaLDA propose également des formats d'export HTML, PDF et RTF, et sa structure XML avec DTD offre une certaine souplesse d'adaptation. En revanche, PRAAT est un outil plus spécialisé et relativement fermé en termes d'échange de données.

## Stabilité, limitations et communauté d'utilisateurs

Du point de vue de la stabilité, certains outils comme ANVIL ou EXMARaLDA sont signalés comme sujets à des arrêts fréquents ou à des contraintes de taille de fichier (ex. : vidéo <10 min). Ce sont des limites notables dans le cadre de l'annotation de longues séquences d'enseignement. À l'inverse, des outils comme PRAAT ou ELAN sont réputés stables et largement utilisés dans la communauté scientifique, ce qui garantit un meilleur support technique, une documentation abondante, et des ressources mutualisées (tutoriels, bibliothèques de scripts…).

## Conclusion

Aucun outil ne répond de manière exhaustive à toutes les exigences de l'analyse des séquences explicatives multimodales. Le choix de l'outil dépend étroitement de la nature des données à traiter, des objectifs d'analyse (linguistique, interactionnelle, gestuelle, prosodique…) et du niveau d'expertise technique du chercheur.

- ELAN s'impose comme l'outil le plus polyvalent et interopérable pour des analyses fines d'interactions filmées.
- EXMARaLDA est particulièrement pertinent pour les projets souhaitant produire des documents lisibles (PDF, HTML) tout en conservant un haut niveau d'annotation.
- CLAN et PRAAT sont à privilégier pour des analyses ciblées, respectivement conversationnelles et acoustiques.
- TRANSANA, bien que payant, propose des fonctionnalités puissantes pour l'analyse qualitative de corpus volumineux.
- ANVIL, enfin, reste un outil de niche, utile notamment pour sa gestion hiérarchique des catégories, mais limité dans la prise en charge de certains formats et peu évolutif.

L'avenir de l'analyse multimodale automatisée repose ainsi non sur un outil unique, mais sur l'articulation raisonnée de plusieurs environnements logiciels, en fonction du corpus, des finalités de recherche et des logiques d'interopérabilité.

# Essais de transcription automatique d'une séquence explicative ou collaborative

Aurélien NGUYEN
TECHNE (6316)
Julie RANCON
(FORELLIS EA3816)

Depuis une dizaine d'années, la transcription automatique de la parole connaît des avancées spectaculaires, portées par les progrès de l'apprentissage profond, la diversification des corpus d'entraînement, et la montée en puissance des interfaces numériques intégrées. Ces avancées s'inscrivent dans un double mouvement. D'un côté, l'accélération des besoins en traitement massif de données orales, notamment dans les sciences humaines et sociales. De l'autre, la nécessité de produire des représentations fiables et exploitables du discours pour alimenter des analyses linguistiques, interactionnelles ou didactiques. Dans ce contexte, la transcription manuelle reste largement prédominante dans les recherches en linguistique de l'oral et en analyse de l'interaction. Elle est reconnue pour sa capacité à restituer la dimension multimodale, contextuelle et sémiotique de l'activité langagière. Mais elle constitue également un goulot d'étranglement méthodologique lente, coûteuse, subjective, elle peut limiter la constitution de grands corpus, ralentir les campagnes d'annotation, et freiner la mutualisation des ressources. C'est dans cette tension entre précision attendue et gain d'automatisation que s'inscrit l'expérimentation conjointe menée par TECHNE et FORELLIS, visant à tester les possibilités offertes par des outils de transcription automatique dans le traitement de séquences pédagogiques réelles. L'objectif n'était pas de déléguer entièrement le travail de transcription, mais d'évaluer la faisabilité d'un modèle hybride associant reconnaissance vocale automatisée et réinterprétation humaine, dans une visée de recherche qualitative sur les interactions en contexte éducatif.

## Présentation de l'expérimentation

### A. Nature des données

Deux types de corpus ont été mobilisés :

1. Un extrait de table ronde en studio, en lien avec l'éducation numérique, issu d'un enregistrement radiophonique de qualité optimale (environnement sonore contrôlé, locuteurs identifiés, discours structuré).
2. Un segment de classe en situation d'enseignement du FLE, capté via un téléphone mobile dans un environnement semi-bruyant, sans micro-cravate, et incluant des tours de parole spontanés, des gestes co-verbaux et des reformulations.

Ces deux extraits présentent des caractéristiques distinctes, représentatives des écarts entre discours institutionnalisé et oral pédagogique ordinaire. Le premier se rapproche d'un genre radiophonique semi-formel, tandis que le second engage une interaction didactique non préparée, avec ses caractéristiques propres : pauses, chevauchements, prosodie emphatique, segments autocorrectifs, adressage co-énonciatif, etc.

### B. Outils testés

Deux dispositifs de transcription automatique ont été mobilisés :

- VOCAPIA, moteur professionnel de reconnaissance vocale développé par la société VoxSigma, réputé pour sa performance dans des contextes médiatiques et multilingues. Il repose sur des modèles acoustiques et linguistiques entraînés à partir de larges corpus institutionnels.
- Dictée vocale iPhone (iOS), système intégré aux appareils Apple, mobilisé ici dans une approche « grand public », sans paramétrage expert, dans une logique de captation légère. Le fichier audio a été dicté à l'outil par une voix humaine imitant l'énoncé réel, puis exporté sous forme textuelle brute avant nettoyage.

### C. Analyse des résultats obtenus

**Transcription VOCAPIA : segmentation lexicale sans structuration interactionnelle**

Le traitement de la séquence institutionnelle par VOCAPIA donne lieu à une transcription lexicalement riche, dans laquelle la majorité des mots prononcés sont correctement reconnus, y compris les mots techniques et les noms propres. Cette performance est à mettre au crédit de la qualité acoustique de l'enregistrement, de l'absence de bruits parasites, et d'un débit relativement contrôlé. Néanmoins, plusieurs limitations structurelles apparaissent. D'abord, la segmentation phrastique est défaillante. Les phrases s'enchaînent sans ponctuation pertinente, les incises sont mal balisées, et les marqueurs d'organisation discursive (connecteurs, reformulations, requalifications) ne sont pas identifiés comme tels. Ensuite, les attributions de parole sont absentes, ce qui neutralise la dimension dialogique du discours. Enfin, certaines expressions sont déformées par le moteur, parfois de manière absurde, comme dans :

« Thierry Pasquier de l'espacement des souffrances au bonjour »

au lieu de :

« Thierry Pasquier de l'Espace Mendès France, bonjour »

Cette erreur, qui relève d'une confusion homophonique sémantiquement aberrante, souligne les limites des modèles de transcription lorsqu'ils ne disposent pas de repères contextuels ou de connaissances de domaine suffisantes. En somme, la transcription produite par VOCAPIA offre une base textuelle utile pour repérer des thématiques générales ou extraire des entités nommées, mais reste inexploitée pour toute analyse de la dynamique interactionnelle, de la prosodie ou de la multimodalité.

**Transcription iPhone : restitution partielle du flux oral, nettoyage nécessaire**

Dans le second cas, la dictée automatique iPhone produit une chaîne verbale continue, sans ponctuation, ni segmentation, ni distinction des locuteurs. Le texte initial obtenu est difficilement lisible, linéaire et saturé d'imprécisions. Un extrait brut illustre ce phénomène :

« résidence secondaireQu'est-ce que ça veut dire c'est la deuxième maison à côté de à côté de la mère la plage ouais c'est la deuxième maison… »

Plusieurs problèmes apparaissent : non-détection des frontières phrastiques, confusion entre homophones (« mère » au lieu de « mer »), absence de pauses syntaxiques, et effacement des fonctions discursives (question, validation, appui…). Cependant, après nettoyage manuel, le texte devient exploitable. Le transcripteur re-segmente les énoncés, identifie les locuteurs, restaure la ponctuation, et reformule certains segments pour en préserver la cohérence. Le passage suivant, une fois édité, devient :

Enseignante : Qu'est-ce que ça veut dire résidence secondaire ?
Étudiante : C'est la deuxième maison, à côté, euh… à côté de la mer, la plage.
Enseignante : Ouais, c'est la deuxième maison ! Beaucoup de gens sont propriétaires de deux maisons : la maison principale et la maison de vacances…

Ce travail de recontextualisation repose sur une connaissance fine de la scène enregistrée et de ses enjeux pédagogiques. On observe ici que l'outil automatique, bien que défaillant en autonomie, peut servir de support initial pour accélérer le travail de transcription, à condition d'être retravaillé dans une optique interprétative.

## Discussion : enjeux méthodologiques et perspectives

Les résultats obtenus montrent clairement que les outils de transcription automatique actuels ne peuvent pas, à ce jour, se substituer au travail de transcription manuelle dans une perspective qualitative, notamment dans l'analyse des interactions en classe. Plusieurs limites persistent : l'impossibilité de détecter les gestes, regards, chevauchements, la neutralisation de la multimodalité et surtout l'incapacité à segmenter de manière adéquate les séquences en unités pragmatiques ou en actes de langage. Néanmoins, ces outils peuvent jouer un rôle facilitateur, notamment pour amorcer une transcription, repérer des unités lexicales saillantes, baliser un enregistrement long, ou générer une première version brute avant annotation. Leur usage dans une logique semi-automatisée, associée à une plateforme d'annotation multimodale (ELAN, Transana, Anvil), pourrait contribuer à réduire le coût cognitif et temporel du travail de transcription, sans en compromettre la rigueur. La comparaison entre les deux outils testés montre aussi que la qualité du signal sonore et la nature du discours influencent fortement la performance du moteur de reconnaissance. Un discours structuré, avec peu de locuteurs, dans un environnement calme, sera bien mieux transcrit qu'un échange en classe, marqué par l'informalité, les hésitations, les reformulations, et les perturbations sonores. Enfin, la dimension interactionnelle et pédagogique du discours oral appelle à penser des modèles de transcription intégrant la diversité des modalités d'expression. Il ne s'agit pas seulement de transcrire des mots, mais de rendre compte d'une activité conjointe, située, gestualisée, ancrée dans un contexte d'apprentissage.

## Conclusion

Les essais menés confirment que la transcription automatique, malgré ses avancées, ne saurait être utilisée en l'état pour des analyses interactionnelles et didactiques complexes. Toutefois, dans une perspective de traitement outillé, elle constitue un point d'appui stratégique pour amorcer ou baliser le travail de transcription. L'avenir réside sans doute dans le développement d'outils hybrides, intelligents, interactifs, permettant de combiner les performances des moteurs de reconnaissance vocale avec les capacités interprétatives des chercheurs. Cela suppose de co-construire des outils avec les chercheurs, d'entraîner les modèles sur des corpus pédagogiques réels, et de penser des interfaces ergonomiques qui rendent possible une annotation semi-automatisée de l'oral multimodal. À ce titre, les résultats obtenus s'inscrivent dans les perspectives ouvertes par les projets BodySMART (Bianchini, 2025) et LexiKHuM (Bianchini, 2023), qui visent précisément à articuler modélisation gestuelle, reconnaissance automatique et typographie annotée. En somme, il ne s’agit plus de demander si l’automatisation peut remplacer la transcription manuelle, mais plutôt comment la technologie peut devenir un partenaire méthodologique du chercheur en SHS, au service d’une analyse du langage aussi rigoureuse que sensible aux pratiques concrètes de l’interaction.

## Essais de traitement automatique de la parole : deux trnascriptions

### Transcription VOCAPIA – son de qualité

[MS1]
Mais le sujet central d'aujourd'hui, ça va être un événement qui va arriver début octobre à

Poitiers, c'est le Campus il éducation le ses 2 E et aujourd'hui pour en parler avec nous, Sylvie Merle fortin de techniques bonjour bonjour et Thierry Pasquier de l'espacement des souffrances au bonjour à tous les 2, alors le campus si Éducation enseignement anciennement plus européen des été ça fait quelques années que ça a changé déjà, donc je pense que dans les esprits, maintenant c'est clair, mais néanmoins terminologie toujours bonne à redéfinir que c'est ce que c'est que l'e-éducation.

[FS2]
Alors l'idée du question c'est l'éducation qui est permise grâce à la multiplication d'et des outils numériques, la diffusion à grande échelle, l'accès à l'information et on l'espère, reconnaissance Un pour tous, à travers les espaces etc et et d'où notre thème, cette année, en lien avec la francophonie puisque justement on espère vraiment que le les techniques numériques nous permettant de, de, d'aller sur tous ces espaces francophones répartis sur les 5 continents.

[MS1]
Oui effectivement, alors là la le la thématique de de cette année va être va être extrêmement prégnante sur sur l'e-éducation c'est une une spécificité à Poitiers, d'avoir autant de de de billes sur ce volet-là.

[FS2]
Oui alors à Poitiers, on a, on a la chance d'avoir un des partenariats extrêmement multiple,

on a le S Pen à cobalt qui regroupe qui un groupement ce qu'on appelle un un cluster de 200 entreprises acteurs en lien avec le numérique et donc le 7 E la Campus il éducation réunit les principaux acteurs privés et puis.

### Essais de traitement automatique de la parole par Iphone

Donc il va manquer ça qu’est-ce que ça veut dire résidence secondaireQu’est-ce que ça veut dire c’est la deuxième maison à côté de à côté de la mère la plage ouais c’est la deuxième maison beaucoup de gens sont propriétaire de deux maisons la maison principale est la maison de vacances la maison secondaire la deuxième pour les vacances et ça peut être n’importe où à à la mer à la montagne à la campagne dans une autre ville ça peut être aussi un appartement dans une grande ville pour les gens qui aiment faire comme Fatiha du shopping et si vous aimez plus la tranquillité vous allez plutôt dans le sud de la Vienne au Bordeleau maison secondaire donc faudra chercher ça ici à quel est le dernier type d'hébergement qui est cité dans le document

**Nettoyage de la dictée automatique par Iphone par Julie Rançon**

Enseignante : Donc il va manquer ça ! Qu'est-ce que ça veut dire résidence secondaire ? Qu'est-ce que ça veut dire.

Étudiante : C'est la deuxième maison à côté euh à côté de la mère la plage

Enseignante : Ouais c'est la deuxième maison ! Beaucoup de gens sont propriétaires de deux maisons : la maison principale et la maison de vacances. La maison secondaire, la deuxième, pour les vacances. Et ça peut être n'importe où ! à la mer, à la montagne, à la campagne, dans une autre ville… Ca peut être aussi un appartement dans une grande ville, pour les gens qui aiment faire comme Fatia du shopping ! Et si vous aimez plus la tranquillité, vous allez plutôt dans le sud de la Vienne, au bord de l'eau. Maison secondaire ! Donc faudra chercher ça ! Ici ! Quel est le dernier type d'hébergement qui est cité dans le document ?

# Le TechnéLAB : une réponse aux difficultés de captations des informations multimodales

Julie Rançon
FORELLIS (EA 3816)
Jean-François Cerisier
TECHNE (EA 6316)

---

Le TechnéLab, salle expérimentale du laboratoire TECHNE (EA 6316) de l'Université de Poitiers, est un dispositif scientifique conçu pour l'étude fine des interactions en contexte éducatif instrumenté. Il s'agit d'un espace dédié à la captation multimodale de haute qualité, équipé de caméras tourelles motorisées, de micros individuels à haute sensibilité, de dispositifs de suivi du regard et de matériels numériques interactifs, permettant l'enregistrement synchronisé des dimensions verbales, paraverbales et non verbales de l'activité pédagogique. Cette salle est conçue à la fois comme environnement d'expérimentation contrôlée et dispositif d'observation instrumentée in situ, avec possibilité de régie distante et de visualisation différée.

Dans le cadre du projet TRAIMA (TRaitement Automatique des Interactions Multimodales en Apprentissage), nous avons mobilisé le TechnéLab pour constituer des corpus pédagogiques multimodaux de haute qualité, destinés à l'étude des ajustements interactionnels dans des situations de transmission du français langue maternelle ou étrangère. Chaque participant – enseignants comme apprenants – devait être équipé d'un micro-cravate individuel, ce qui aurait dû permettre de capter la parole avec une précision acoustique remarquable, indépendamment de la distance au micro ambiant ou du niveau sonore ambiant.

Toutefois, cette campagne expérimentale a été brutalement interrompue par le confinement national du mois de mars 2020, lié à la pandémie de COVID-19. Cette suspension a limité l'ampleur du corpus recueilli et a empêché, à ce stade, la systématisation des enregistrements prévus initialement. Elle a également mis en lumière la fragilité logistique des protocoles expérimentaux dépendant d'un espace physique commun, et a renforcé, en retour, la nécessité d'articuler les approches de terrain à des méthodes d'annotation à distance et de traitement semi-automatisé des données multimodales.

# Le traitement automatique des données multimodales :

## Le cas du discours explicatif

Julie RANCON
FORELLIS (EA 3816)

---

Le projet TRAIMA a permis d'ouvrir une voie féconde à l'interface des sciences du langage, des sciences de l'éducation et des sciences du numérique, en explorant les conditions techniques, méthodologiques et théoriques d'un traitement automatique des interactions multimodales en contexte pédagogique. En mobilisant un corpus vidéo finement instrumenté – capté au sein du TechnéLab à l'aide de dispositifs audio-visuels de haute précision – nous avons pu isoler et annoter des séquences didactiques représentatives, où se jouent les dynamiques complexes d'ajustement, de reformulation, de co-construction du sens, et d'engagement corporel des acteurs. L'analyse manuelle et assistée de ces données a mis en lumière à la fois les potentialités et les limites actuelles des outils de transcription automatique appliqués à des interactions naturelles, en particulier dans des environnements bruités, riches en chevauchements, gestes, et variations prosodiques. Nos expérimentations montrent que si les moteurs de reconnaissance vocale de dernière génération (comme Vocapia ou les modules de dictée vocale mobiles) permettent désormais une captation fluide du signal verbal, leur usage reste encore insuffisant pour l'annotation multimodale fine, qui suppose une intégration synchronisée des canaux gestuels, prosodiques et posturaux. Dès lors, l'enjeu n'est pas de substituer le traitement automatique au travail du chercheur, mais bien d'en faire un vecteur de gain méthodologique en réduisant le temps de segmentation initiale, en proposant des alignements pré-annotés, et en facilitant les croisements avec des typologies gestuelles ou interactionnelles. L'horizon visé est celui d'un traitement semi-automatisé et outillé, dans lequel l'intelligence humaine et l'intelligence computationnelle s'articulent dans un processus coopératif.

Les travaux amorcés dans le cadre de TRAIMA posent ainsi les jalons d'une linguistique de corpus multimodale augmentée, sensible aux complexités du terrain éducatif, mais appuyée sur des standards de qualité et d'interopérabilité scientifique. En l'état, le corpus collecté demeure un matériau unique, exploitable par la communauté scientifique pour l'étude des interactions verbales et non verbales en classe. Il appelle à être enrichi, mutualisé et prolongé par de futurs projets, qui pourraient intégrer des dimensions d'analyse émotionnelle, de traitement visuel automatisé des gestes, ou encore de modélisation des profils interactionnels enseignants-apprenants.

Ce projet invite enfin à réfléchir plus largement à la place des corpus ouverts, éthiques et traçables dans la recherche en SHS à l'ère des humanités numériques : une exigence scientifique autant qu'un engagement politique en faveur d'une science reproductible, transparente et collaborative.